\documentclass[11pt]{article}

\usepackage[preprint]{acl}

\usepackage{times}
\usepackage{latexsym}

\usepackage[T1]{fontenc}

\usepackage[utf8]{inputenc}
\usepackage{listings}
\usepackage{microtype}

\newcommand{\gain}[1]{{\color{teal}\scriptsize{($\uparrow$#1)}}}
\newcommand{\drop}[1]{{\color{purple}\scriptsize{($\downarrow$#1)}}}
\newcommand{\neut}[1]{{\color{gray}\scriptsize{($+$#1)}}}
\newcommand{\nfe}[1]{{\color{gray}#1}}

\usepackage{inconsolata}

\usepackage{graphicx}
\usepackage{enumitem}

\usepackage[utf8]{inputenc}
\usepackage[T1]{fontenc}
\usepackage{microtype}              

\usepackage{amsmath}                
\usepackage{amssymb}                
\usepackage{amsthm}                 
\usepackage{mathtools}              
\usepackage{bm}                     

\usepackage{algorithm}              
\usepackage{algpseudocode}          

\usepackage{graphicx}
\usepackage{booktabs}               
\usepackage{array}
\usepackage{multirow}
\usepackage{subcaption}
\usepackage{placeins}
\usepackage{hyperref}               
\usepackage{cleveref}               

\theoremstyle{plain}

\theoremstyle{definition}

\theoremstyle{remark}

\hypersetup{
    colorlinks=true,
    linkcolor=blue!60!black,
    citecolor=blue!60!black,
    urlcolor=blue!60!black,
}

\title{
CoDR: Training-Free Confidence-Drift Remasking \\
for Diffusion Language Models
}

\author{
  \textbf{Yue Wu}\textsuperscript{1,\textdagger},
  \textbf{Qinghe Zheng}\textsuperscript{2,\textdagger},
  \textbf{Yu Zhang}\textsuperscript{2,*},
  \textbf{Jian Huang}\textsuperscript{1,*}
  \\
  \textsuperscript{1}The Hong Kong Polytechnic University, Hong Kong, China
  \\
  \textsuperscript{2}Southern University of Science and Technology, Shenzhen, China
  \\
  \small{
    \texttt{yue0301.wu@connect.polyu.hk},
    \texttt{12532603@mail.sustech.edu.cn}
  }
  \\
  \small{
    \texttt{yu.zhang.ust@gmail.com},
    \texttt{jianhuang@polyu.edu.hk}
  }
  \\
  \small{
    \textsuperscript{\textdagger}Equal contribution.
    \textsuperscript{*}Corresponding authors.
  }
}

\begin{document}
\maketitle
\begin{abstract}
Masked diffusion language models (MDLMs) decode by repeatedly committing tokens to masked positions, but these commitments are usually irreversible. A token chosen under sparse, partial context is kept fixed, even when later context no longer supports it. Existing samplers mainly decide when to commit a token, but rarely check whether an already committed token should still be kept, allowing early mistakes to propagate. We trace this issue to confidence drift, where the model's confidence in a committed token drops from its sparse commit-time context to the denser context available later. Based on this signal, we propose \textbf{CoDR} (\textbf{Co}nfidence-\textbf{D}rift \textbf{R}emasking), a training-free and sampler-agnostic refinement pass. CoDR estimates drift for all committed positions in only $k$ forward passes via $k$-partition probing, then remasks and regenerates only the tokens the model no longer endorses. Across two backbones, four reasoning and coding tasks, and three base samplers, CoDR improves average accuracy across all evaluated model–sampler configurations and improves most individual task settings with modest overhead. Controlled experiments show that the gains come from targeted confidence-drift remasking rather than extra compute alone, and that CoDR uses far fewer forward passes than prior remasking methods. Code is available at \url{https://github.com/YueWu0301/CoDR}.
\end{abstract}

\section{Introduction}
\label{sec:intro}
Masked diffusion language models (MDLMs) \cite{austin2021d3pm} provide a parallel alternative to autoregressive generation~\citep{sahoo2024mdlm,nie2025llada,ye2025dream}. Starting from a fully masked sequence, they iteratively commit tokens until generation is complete, enabling flexible any-order decoding~\citep{ou2024radd,kim2025train} and potentially lower inference latency. With model parameters fixed at inference time, recent work has therefore focused on samplers that improve commitment quality and efficiency~\citep{wu2025fastdllm,benhamu2025ebsampler}. However, these methods rarely revisit tokens after they are committed.

\begin{figure}[H]
\centering
\includegraphics[width=0.4\textwidth]{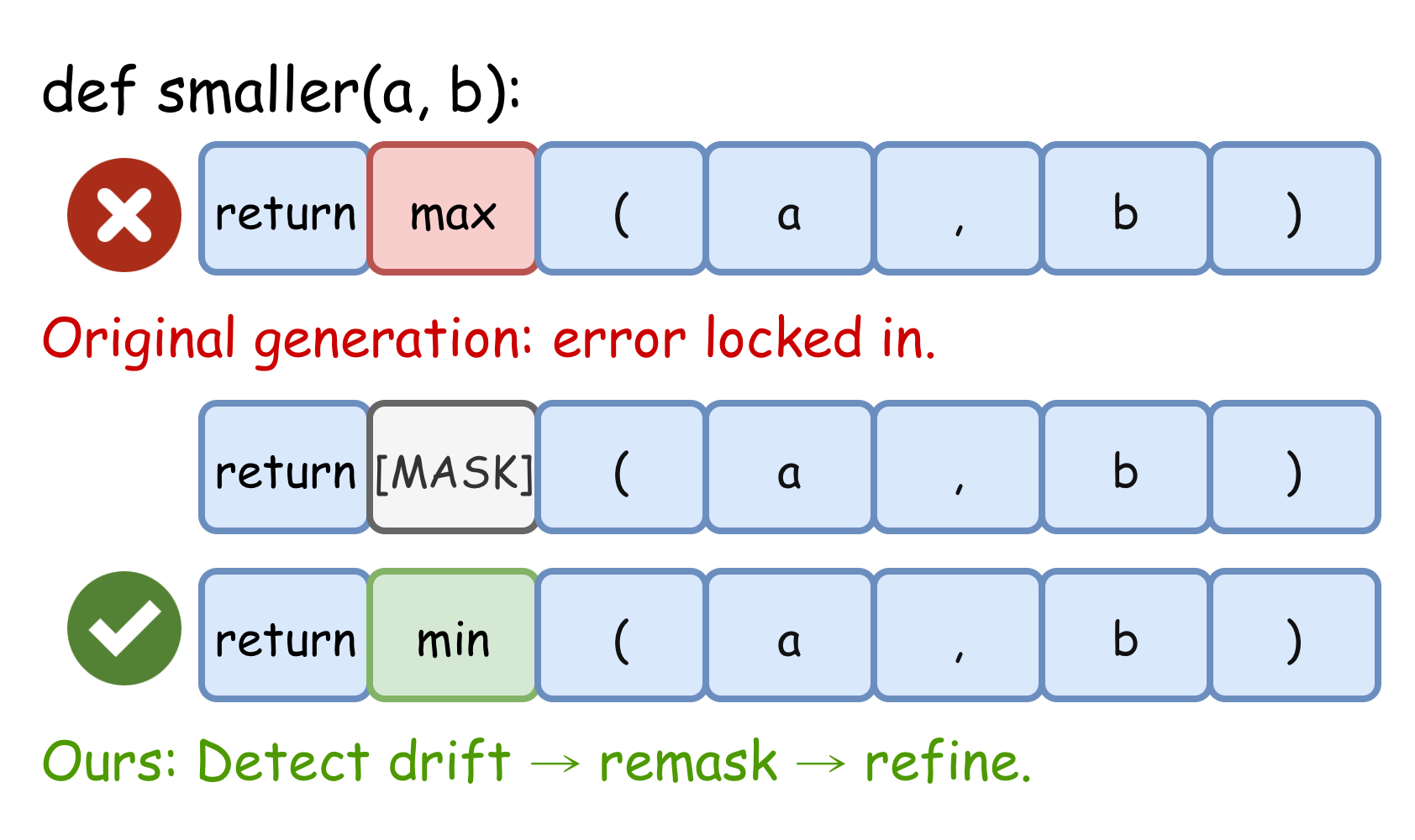}
\caption{Standard generation fixes tokens once committed, leaving early mistakes unchanged. Drift-guided refinement detects unstable committed tokens, remasks them, and refills them under the updated context.
}
\label{fig:figure1}
\end{figure}

This creates irreversible commitment: a token selected under incomplete context remains fixed even if later context no longer supports it. A locally plausible token at commitment time may become incompatible after surrounding positions are filled. The problem is amplified when several positions are committed in parallel, because co-committed tokens do not condition on one another and may introduce local inconsistencies~\citep{gu2018nat,wu2025fastdllm,wu2026swd}. These early commitments then condition later predictions, allowing errors to propagate through subsequent decoding. A natural remedy is to revisit unstable tokens once richer context becomes available rather than treating every commitment as final.

Existing samplers use confidence, entropy, or attention signals to decide which masked positions to commit~\citep{wu2025fastdllm,benhamu2025ebsampler,zhou2026dos}, making the initial commitment more cautious but usually keeping committed tokens fixed. Recent remasking methods relax this constraint: ReMDM iteratively revisits commitments~\citep{wang2025remdm}, while CORE probes robustness to perturbed contexts~\citep{zhai2026core}. These approaches establish the value of reversible generation, but do not directly measure how support for a token changes as decoding progresses. We instead ask whether the model's confidence in the same token decreases from its early commit-time context to a later, denser context. Such confidence drift directly identifies commitments that lose model support during decoding.

We propose \textbf{CoDR} (\textbf{Co}nfidence-\textbf{D}rift \textbf{R}emasking), a training-free refinement pass (Figure~\ref{fig:figure1}). CoDR compares each token's commit-time confidence with its confidence under later context and treats a confidence drop as evidence that the updated context provides less support. To avoid one forward pass per position, $k$-partition probing estimates drift for all commitments in only $k$ passes. CoDR then remasks the selected positions and invokes the original sampler to refill them under denser bidirectional context, while leaving all other tokens fixed. It therefore acts as a sampler-agnostic plug-in for decoders that expose commit-time probabilities and support masked refill. Across LLaDA-8B~\citep{nie2025llada} and Dream-7B~\citep{ye2025dream}, multiple samplers, and reasoning and coding tasks, CoDR improves the accuracy--compute trade-off in most settings. Controlled comparisons attribute these gains to targeted confidence-drift revision rather than random remasking or extra decoding steps alone.

Our contributions are summarized as follows:
\begin{itemize}[leftmargin=*,topsep=2pt,itemsep=1pt]
    \item We identify confidence drift as a temporal symptom of decoding instability, where committed tokens lose model support as context grows.
    \item We propose CoDR, a training-free plug-in that efficiently detects and revises drifted commitments without modifying the base model or sampler.
    \item We show across models, tasks, and samplers that targeted revision improves the accuracy--compute trade-off over unguided extra computation.
\end{itemize}

\begin{figure*}[t]
\centering
\includegraphics[width=0.8\textwidth]{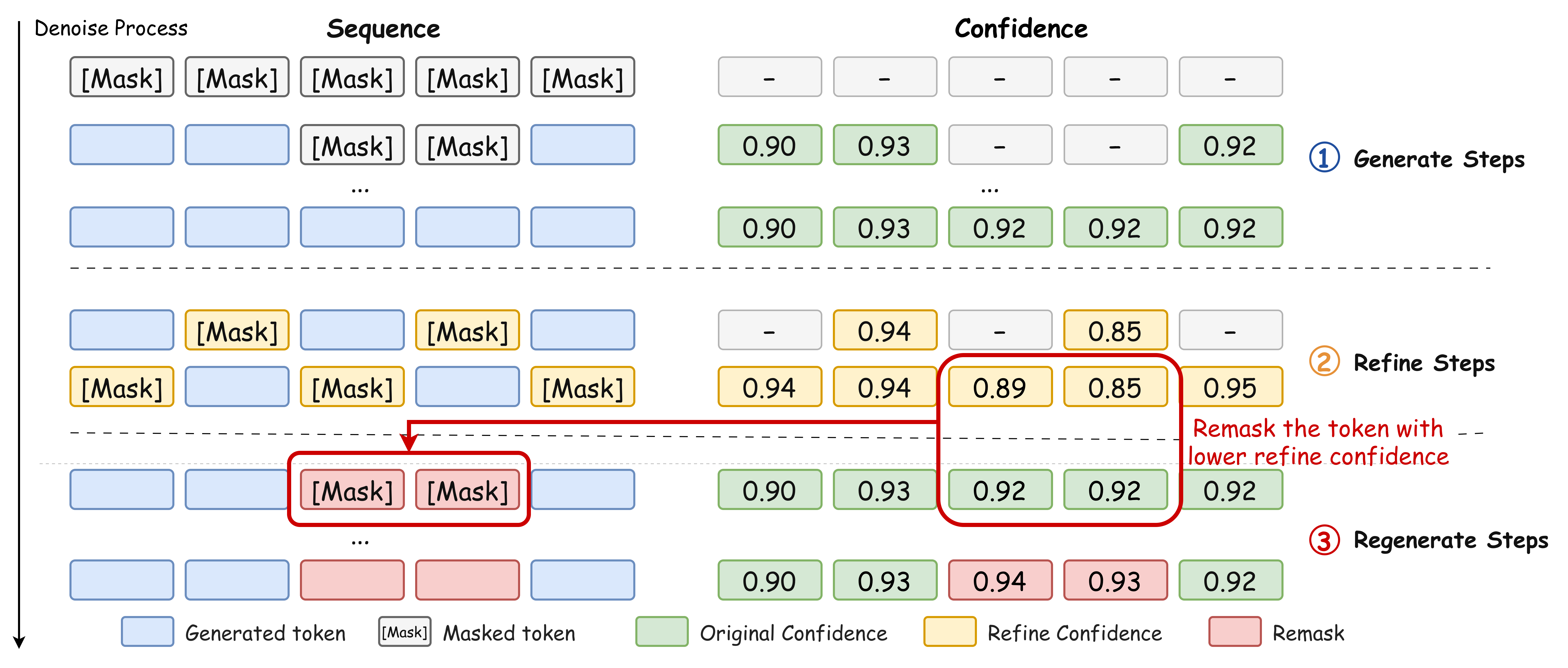}
\caption{Overview of our method CoDR. The base sampler first fills a block, then the refine pass re-scores committed tokens under richer context, remasks positions whose confidence drifts downward, and refills only those positions.}
\label{fig:block-wise-refinement}
\end{figure*}

\section{Related Work}
\label{sec:related}

\paragraph{Masked Diffusion Language Models.}
Masked diffusion language modeling grew out of discrete diffusion.
D3PM~\citep{austin2021d3pm} introduced discrete-state diffusion and showed that an absorbing ``mask'' state is particularly natural for language; subsequent work such as SEDD~\citep{lou2024sedd}, MDLM~\citep{sahoo2024mdlm}, and RADD~\citep{ou2024radd} developed increasingly effective objectives and parameterizations for discrete and masked diffusion language modeling.
Scaling these ideas produced billion-parameter models competitive with autoregressive LLMs. LLaDA~\citep{nie2025llada} demonstrated large-scale pretraining from scratch, while DiffuLLaMA~\citep{gong2024diffullama}, Dream~\citep{ye2025dream}, DiffuCoder~\citep{diffucoder}, and LLaDA 2.0~\citep{llada20} explored adaptation or initialization from pretrained autoregressive models.
The paradigm has further expanded to block-wise generation~\citep{arriola2025bd3lm} and multimodal modeling~\citep{you2025lladav,mmada}.
Recent work has also explored supervised, preference-based, parameter-efficient, and reinforcement-learning-based post-training of diffusion language models~\citep{zhao2025d1scalingreasoningdiffusion,llada15,wang2026nara}, whereas CoDR improves generation without updating model parameters.

\paragraph{Decoding Samplers and Remasking.}
How a trained MDLM is decoded is a second, largely separate question. Existing samplers use model-internal signals such as entropy or margin~\citep{koh2025entropy,kim2025train}, confidence thresholds~\citep{wu2025fastdllm}, entropy bounds~\citep{benhamu2025ebsampler}, denoising history~\citep{wu2026swd,wang2025creditdecoding}, KL criteria~\citep{kim2025klass}, or attention patterns~\citep{zhou2026dos,kim2026dapd} to decide which positions to commit and when. These methods improve the commitment rule, but committed tokens are typically kept fixed. Recent remasking methods relax this constraint by allowing committed tokens to be revisited. ReMDM~\citep{wang2025remdm} introduces an iterative remasking process, while CORE~\citep{zhai2026core} selects revision targets by testing token robustness under perturbed contexts. Our method also belongs to this reversible-generation family, but it uses a different signal: confidence drift, the confidence drop of the same committed token from its commit-time context to a later decoding context. This directly targets tokens whose support decreases as more surrounding context becomes available.

\section{Method}
\label{sec:method}

We introduce \textbf{CoDR} (\textbf{Co}nfidence-\textbf{D}rift \textbf{R}emasking), a training-free refinement method that detects drifted commitments by comparing commit-time and later-context confidence. CoDR estimates this signal with $k$-partition probing, then selectively remasks and refills drifted tokens with the original base sampler (Figure~\ref{fig:block-wise-refinement}).

\subsection{Preliminaries: MDLM Decoding}
\label{sec:prelim}

Let $\mathcal{V}$ be a vocabulary and $\mathtt{[M]} \notin \mathcal{V}$ a special mask token. We decode a target sequence of length $L$, denoted as $x = (x^1, \dots, x^L)$ where each $x^i \in \mathcal{V} \cup \{\mathtt{[M]}\}$, conditioned on a prompt (omitted in notation for brevity). The MDLM is parameterized by a denoising network $f_\theta$ that maps a partially masked sequence to a logit tensor $f_\theta(x) \in \mathbb{R}^{L \times |\mathcal{V}|}$ in a single forward pass. Its per-position predictive distribution over the vocabulary is given by:
\begin{equation}
\label{eq:pred}
p_\theta^{\,i}(\cdot \mid x) = \mathrm{Softmax}\bigl([f_\theta(x)]_i\bigr) \in \Delta(\mathcal{V}).
\end{equation}

Decoding proceeds over discrete steps $t = 1, \dots, T$. Let $C_t \subseteq \{1, \dots, L\}$ denote the set of committed indices and $M_t = \{1, \dots, L\} \setminus C_t$ the set of masked indices after step $t$. Initially, $C_0 = \emptyset$ and $M_0 = \{1, \dots, L\}$, with all positions in $M_0$ holding $\mathtt{[M]}$.

At step $t$, a sampler selects a subset of masked positions $S_t \subseteq M_{t-1}$ to commit. Each token is drawn from its per-position marginal distribution, conditioned on the current partial sequence $x_{t-1}$:

\begin{equation}
\label{eq:unmask}
\hat{x}_t^i \sim p_\theta^{\,i}(\cdot \mid x_{t-1}), \quad \forall i \in S_t.
\end{equation}

The sequence is then updated position-wise as
\begin{equation}
\label{eq:decode-update}
x_t^i =
\begin{cases}
\hat{x}_t^i, & i \in S_t \quad \text{(unmask)},\\
\mathtt{[M]}, & i \in M_{t-1} \setminus S_t \quad \text{(stay masked)},\\
x_{t-1}^i, & i \notin M_{t-1} \quad \text{(already committed)}.
\end{cases}
\end{equation}

The sets are then updated as $C_t = C_{t-1} \cup S_t$ and $M_t = M_{t-1} \setminus S_t$. Committing a group in this manner treats the positions in $S_t$ as conditionally independent given $x_{t-1}$.

Crucially, when position $i$ is committed at step $t$ (i.e., $i \in S_t$), we record its commit-time confidence:
\begin{equation}
\label{eq:psi-old}
\psi^{\mathrm{old}}_i = p_\theta^{\,i}\bigl(\hat{x}_t^i \mid x_{t-1}\bigr).
\end{equation}
This scalar evaluates the marginal probability the network assigned to the chosen token $\hat{x}_t^i$ under the sparse, localized context $x_{t-1}$ available at that exact step. This single recorded scalar per position is the only interface our refinement procedure requires from the underlying base sampler.

\subsection{Confidence-Drift Detection via k-Partition Probing}
\label{sec:drift}

Because each position $i$ is committed at some step $t$ while many positions in $M_{t-1}$ remain masked, $x^i$ is decided under a sparser context than it will eventually reside in. Furthermore, within a co-committed group $S_t$, positions are sampled independently of one another. The optimal token at $i$ can therefore change once more context is filled in at a later time $t' > t$. We define this discrepancy as \emph{confidence drift}, and detect it by re-scoring the model's confidence in the initially chosen token under the denser, later context.

Let $\tilde{x}^{[i \to \mathtt{[M]}]}$ denote the current sequence at time $t'$ with position $i$ artificially re-masked. The re-scored confidence is the predictive probability of the same originally committed token $x^i$ under this richer context, and the drift is its confidence drop from the moment of commitment:
\begin{equation}
\label{eq:drift}
\psi^{\mathrm{new}}_i = p_\theta^{\,i}\bigl(x^i \mid \tilde{x}^{[i \to \mathtt{[M]}]}\bigr), \qquad \delta_i = \psi^{\mathrm{old}}_i - \psi^{\mathrm{new}}_i.
\end{equation}
Both confidences evaluate the probability of the same token $x^i$ via Equation~\ref{eq:pred}. The only difference is the conditioning context: $\psi^{\mathrm{old}}_i$ is computed under the sparse context available at commitment time, whereas $\psi^{\mathrm{new}}_i$ is computed under the denser context available at the current step. A large positive $\delta_i$ indicates that $x^i$ was initially committed with high confidence, but is no longer strongly supported once its surrounding tokens become visible. We treat this confidence drop as the primary signal of a drifted commitment.

Because the logarithm is strictly increasing, the condition $\delta_i>0$ is equivalent to $\log \psi_i^{\mathrm{old}}-\log \psi_i^{\mathrm{new}}>0$. CoDR therefore revisits a commitment precisely when it loses model support under the later, denser context. This does not imply that the token is necessarily incorrect or that confidence drift is a universally optimal revision rule; it is a temporal instability signal whose practical informativeness is supported by the trigger comparisons in Table~\ref{tab:random-trigger} and the revision diagnostics in Section~\ref{sec:rq3}.

Evaluating $\psi^{\mathrm{new}}_i$ exactly requires one forward pass per committed position, resulting in $|C|$ passes per invocation. This overhead is prohibitively expensive for generation. Instead, we approximate $\psi^{\mathrm{new}}_i$ for all committed positions simultaneously in just $k$ passes. We partition the committed set $C$ into $k$ residue classes by stride $k$, defining $C^{(r)} = \{i \in C : i \equiv r \!\!\pmod{k}\}$ for $r = 0, \dots, k-1$. For each class, we mask the positions as a whole:
\begin{equation}
\label{eq:probe}
\tilde{x}^{j, (r)} = \begin{cases} \mathtt{[M]} & j \in C^{(r)}, \\ x^j & \text{otherwise.} \end{cases}
\end{equation}
A single forward pass on $\tilde{x}^{(r)}$ yields $\psi^{\mathrm{new}}_i \gets p_\theta^{\,i}(x^i \mid \tilde{x}^{(r)})$ for every $i \in C^{(r)}$. Thus, $k$ passes assign a $\delta_i$ to all committed positions. This probe is exact when $k = |C|$ (where each class is a singleton); for smaller $k$, each $i$ is re-scored while its same-class neighbors are also masked. The stride-$k$ spacing distributes masked positions evenly across the sequence, helping preserve regular local context around each probed position. We use $k = 2$ by default, separating even and odd indices. We discuss the importance of confidence-drift and k-partition probing in Experiment \ref{sec:rq2} and its ablation study in Experiment \ref{sec:ablation}.
\begin{algorithm}[t]
\caption{CoDR decoding with confidence-drift remasking}
\label{alg:main}
\begin{algorithmic}[1]
\Require MDLM $f_\theta$, sequence length $L$, block size $B$, drift threshold $\tau_{\mathrm{drift}}$, partition count $k$
\State $x \gets \mathtt{[M]}^{L}$; \quad $C \gets \emptyset$; \quad $\mathcal{R} \gets \emptyset$

\For{$b = 1$ \textbf{to} $\lceil L/B \rceil$}
    \State $\mathcal{B}_b \gets \{i : (b{-}1)B < i \leq \min(bB,L)\}$
    \State $(x, C, \psi^{\mathrm{old}}) \gets \mathrm{Sampler}(\mathcal{B}_b)$
    \Comment{decode block $\mathcal{B}_b$}

    \For{$r = 0$ \textbf{to} $k-1$}
        \State $C^{(r)} \gets \{i \in C : i \equiv r \pmod{k}\}$
        \State $\tilde{x} \gets x$; \quad $\tilde{x}^{i} \gets \mathtt{[M]} \ \forall i \in C^{(r)}$
        \State $\psi^{\mathrm{new}}_i \gets p_\theta^{\,i}(x^i \mid \tilde{x})
        \quad \forall i \in C^{(r)}$
    \EndFor

    \State $\mathcal{R} \gets
    \{i \in C : \psi^{\mathrm{old}}_i - \psi^{\mathrm{new}}_i > \tau_{\mathrm{drift}}\}$

    \If{$\mathcal{R} \neq \emptyset$}
        \State $x^i \gets \mathtt{[M]} \quad \forall i \in \mathcal{R}$; \quad $C \gets C \setminus \mathcal{R}$
        \State $(x, C, \psi^{\mathrm{old}}) \gets \mathrm{Sampler}(\mathcal{R})$
\Comment{fill the re-masked positions}
    \EndIf
\EndFor

\State \Return $x$
\end{algorithmic}
\end{algorithm}

\begin{table*}[t]
\centering
\small
\caption{Comprehensive benchmark results over four tasks. \textbf{Acc. (\%)} is accuracy, and \textbf{NFE} is the average number of forward evaluations. Parenthesized annotations indicate absolute accuracy-point changes brought by refinement. \textbf{Average} reports the unweighted mean Acc. and NFE across the four tasks. }
\label{tab:main}
\setlength{\tabcolsep}{2pt} 
\begin{tabular}{l l l l l l l l l l l l}
\toprule
\multirow{2}{*}{Model} & \multirow{2}{*}{Sampler} & \multicolumn{2}{c}{HumanEval} & \multicolumn{2}{c}{MBPP} & \multicolumn{2}{c}{MATH-500} & \multicolumn{2}{c}{GSM8K} & \multicolumn{2}{c}{\textbf{Average}} \\
\cmidrule(lr){3-4} \cmidrule(lr){5-6} \cmidrule(lr){7-8} \cmidrule(lr){9-10} \cmidrule(lr){11-12}
 & & \multicolumn{1}{c}{Acc. (\%)} & \multicolumn{1}{c}{NFE}
   & \multicolumn{1}{c}{Acc. (\%)} & \multicolumn{1}{c}{NFE}
   & \multicolumn{1}{c}{Acc. (\%)} & \multicolumn{1}{c}{NFE}
   & \multicolumn{1}{c}{Acc. (\%)} & \multicolumn{1}{c}{NFE}
   & \multicolumn{1}{c}{\textbf{Acc. (\%)}} & \multicolumn{1}{c}{NFE} \\
\midrule
\multirow{7}{*}{LLaDA}
 & top1                     & $41.46$ & \nfe{$256$} & $39.20$ & \nfe{$256$} & $38.60$ & \nfe{$256$} & $72.10$ & \nfe{$256$} & $47.84$ & \nfe{$256$} \\
 & \quad+CoDR             & $42.68$ \gain{1.22} & \nfe{$304$} & $\bm{40.60}$ \gain{1.40} & \nfe{$312$} & $38.40$ & \nfe{$314$} & $73.09$ \gain{0.99} & \nfe{$304$} & $48.69$ \gain{0.85} & \nfe{$309$} \\
\addlinespace[2pt]
 & fd $(\tau{=}0.9)$        & $42.68$ & \nfe{$\phantom{0}79$} & $39.00$ & \nfe{$\phantom{0}84$} & $36.80$ & \nfe{$105$} & $73.01$ & \nfe{$\phantom{0}74$} & $47.87$ & \nfe{$\phantom{0}86$} \\
 & \quad+CoDR             & $\bm{43.90}$ \gain{1.22} & \nfe{$113$} & $40.40$ \gain{1.40} & \nfe{$125$} & $37.20$ \gain{0.40} & \nfe{$145$} & $\bm{74.22}$ \gain{1.21} & \nfe{$110$} & $\bm{48.93}$ \gain{1.06} & \nfe{$123$} \\
\addlinespace[2pt]
 & eb $(\gamma{=}0.2)$      & $40.24$ & \nfe{$\phantom{0}80$} & $38.80$ & \nfe{$\phantom{0}82$} & $37.80$ & \nfe{$101$} & $71.27$ & \nfe{$\phantom{0}73$} & $47.03$ & \nfe{$\phantom{0}84$} \\
 & \quad+CoDR             & $41.46$ \gain{1.22} & \nfe{$114$} & $40.20$ \gain{1.40} & \nfe{$125$} & $\bm{39.80}$ \gain{2.00} & \nfe{$143$} & $72.40$ \gain{1.13} & \nfe{$109$} & $48.47$ \gain{1.44} & \nfe{$123$} \\
\addlinespace[4pt]
 & ReMDM                    & $41.46$ & \nfe{$512$} & $40.00$ & \nfe{$512$} & $39.00$ & \nfe{$512$} & $73.31$ & \nfe{$512$} & $48.44$ & \nfe{$512$} \\
 & CORE                     & $38.41$ & \nfe{$272$} & $39.40$ & \nfe{$272$} & $37.40$ & \nfe{$272$} & $$72.25$$& \nfe{$272$}& $46.87$ &\nfe{$272$}  \\

\midrule
\multirow{7}{*}{Dream}
 & top1                     & $57.32$ & \nfe{$256$} & $54.40$ & \nfe{$256$} & $46.60$ & \nfe{$256$} & $80.89$ & \nfe{$256$} & $59.80$ & \nfe{$256$} \\
 & \quad+CoDR             & $\bm{59.15}$ \gain{1.83} & \nfe{$317$} & $54.60$ \gain{0.20} & \nfe{$339$} & $47.60$ \gain{1.00} & \nfe{$320$} & $81.12$ \gain{0.23} & \nfe{$321$} & $\bm{60.62}$ \gain{0.82} & \nfe{$324$} \\
\addlinespace[2pt]
 & fd $(\tau{=}0.9)$        & $54.88$ & \nfe{$\phantom{0}78$} & $54.20$ & \nfe{$105$} & $47.20$ & \nfe{$116$} & $80.06$ & \nfe{$\phantom{0}64$} & $59.09$ & \nfe{$\phantom{0}91$} \\
 & \quad+CoDR             & $57.32$ \gain{2.44} & \nfe{$121$} & $54.40$ \gain{0.20} & \nfe{$158$} & $\bm{48.20}$ \gain{1.00} & \nfe{$158$} & $80.89$ \gain{0.83} & \nfe{$101$} & $60.20$ \gain{1.11} & \nfe{$135$} \\
\addlinespace[2pt]
 & eb $(\gamma{=}0.2)$      & $56.71$ & \nfe{$\phantom{0}77$} & $\bm{54.80}$ & \nfe{$100$} & $47.80$ & \nfe{$111$} & $79.98$ & \nfe{$\phantom{0}66$} & $59.82$ & \nfe{$\phantom{0}89$} \\
 & \quad+CoDR             & $\bm{59.15}$ \gain{2.44} & \nfe{$124$} & $\bm{54.80}$ \neut{0.00} & \nfe{$159$} & $47.20$ & \nfe{$155$} & $81.27$ \gain{1.29} & \nfe{$109$} & $60.61$ \gain{0.79} & \nfe{$137$} \\
\addlinespace[4pt]
 & ReMDM                    & $57.93$ & \nfe{$512$} & $54.40$ & \nfe{$512$} & $47.00$ & \nfe{$512$} & $\bm{81.73}$ & \nfe{$512$} & $60.27$ & \nfe{$512$} \\
 & CORE                     & $57.32$ & \nfe{$272$} & $53.80$ & \nfe{$272$} & $45.00$ & \nfe{$272$} & $79.61$& \nfe{$272$}& $58.93$ &\nfe{$272$}  \\
\bottomrule
\end{tabular}
\end{table*}

\subsection{Confidence-Drift Remasking and Refill}
\label{sec:refill_and_alg}

Given the confidence-drift scores $\{\delta_i\}_{i\in C}$, CoDR intervenes only on commitments whose support has decreased beyond a tolerance threshold $\tau_{\mathrm{drift}}$. The remask set is defined as
\begin{equation}
\label{eq:remask}
\mathcal{R} = \{\, i \in C \mid \delta_i > \tau_{\mathrm{drift}} \,\}.
\end{equation}

If no drifted position is detected, i.e., $\mathcal{R}=\emptyset$, CoDR keeps the current sequence unchanged. Otherwise, it applies a remask--refill step. Each position in $\mathcal{R}$ is reset to $\mathtt{[M]}$ and removed from the committed set, after which the base sampler is invoked again to fill these re-masked positions. Since later tokens have already been generated, this refill step conditions on a denser bidirectional context than the original commitment. Once the positions are regenerated, the base sampler also refreshes their commit-time confidences $\psi^{\mathrm{old}}_i$, so that future refinement steps compare against the latest commitment.

Algorithm~\ref{alg:main} instantiates this procedure as a post-block method with refinement after every decoded block. At each refinement point, CoDR probes all currently committed positions, including those from previous blocks, and selectively refills the positions in $\mathcal{R}$ whose confidence has drifted. The refinement frequency can be reduced without changing the probe--remask--refill logic; we compare every-block, periodic, final-only, and mid-block schedules in Table~\ref{tab:refinement-schedule}.

\begin{table*}[t]
\centering
\small
\caption{Trigger ablations using \texttt{fd\_refine} on LLaDA-8B ($0$-shot) for HumanEval and GSM8K. The \texttt{bottom-$m$} baselines remask the $m$ committed tokens with the lowest current confidence $\psi^{\mathrm{new}}$, without comparing against their commit-time confidence $\psi^{\mathrm{old}}$.}
\label{tab:random-trigger}
\begin{tabular}{l l l l l l l l l}
\toprule
& \multicolumn{4}{c}{HumanEval} & \multicolumn{4}{c}{GSM8K} \\
\cmidrule(lr){2-5}\cmidrule(lr){6-9}
Trigger & \multicolumn{1}{c}{Acc. (\%)} & \multicolumn{1}{c}{NFE} & \multicolumn{1}{c}{Rem.} & \multicolumn{1}{c}{Edit (\%)}
& \multicolumn{1}{c}{Acc. (\%)} & \multicolumn{1}{c}{NFE} & \multicolumn{1}{c}{Rem.} & \multicolumn{1}{c}{Edit (\%)} \\
\midrule
\texttt{fd} baseline & $42.68$ & $\phantom{0}79$ & $0.0$ & --- & $76.00$ & $\phantom{0}74$ & $0.0$ & --- \\
random $p=0.045$ & $42.68$ \neut{0.00} & $104$ & $50.9$ & $0.21$ & $76.00$ \neut{0.00} & $\phantom{0}99$ & $51.3$ & $0.28$ \\
random $p=0.20$ & $42.68$ \neut{0.00} & $108$ & $227.6$ & $0.27$ & $75.80$ \drop{0.20} & $103$ & $229.7$ & $0.27$ \\
\midrule
bottom $m=6$ & $42.68$ \neut{0.00} & $120.5$ & $48.0$ & 4.83 & $76.80$ \gain{0.80} & $120.2$ & $48.0$ & 5.43 \\
bottom $m=8$ & $43.29$ \gain{0.61} & $123.8$ & $64.0$ & 3.88 & $77.00$ \gain{1.00} & $123.5$ & $64.0$ & 4.31 \\
\midrule
\texttt{top1\_change} & $43.29$ \gain{0.61} & $\phantom{0}99$ & $3.9$ & $36.02$ & $76.80$ \gain{0.80} & $\phantom{0}94$ & $3.8$ & $32.05$ \\
drift (\texttt{conf\_drift}) & $\bm{43.90}$ \gain{1.22} & $113$ & $51.6$ & $3.72$ & $\bm{77.40}$ \gain{1.40} & $110$ & $57.2$ & $3.20$ \\
\bottomrule
\end{tabular}
\end{table*}

\begin{table*}[t]
\centering
\small
\caption{Structured stride vs.\ random partition using \texttt{fd\_refine} on LLaDA-8B ($0$-shot) across probe strides $k$. Random partition results are reported as mean$\pm$std over 5 seeds.}
\label{tab:partition-rand}
\begin{tabular}{c l l l l l l l l l}
\toprule
& & \multicolumn{4}{c}{HumanEval} & \multicolumn{4}{c}{GSM8K} \\
\cmidrule(lr){3-6}\cmidrule(lr){7-10}
\multicolumn{1}{c}{$k$} & Partition
& \multicolumn{1}{c}{Acc. (\%)} & \multicolumn{1}{c}{NFE} & \multicolumn{1}{c}{Rem.} & \multicolumn{1}{c}{Edit (\%)}
& \multicolumn{1}{c}{Acc. (\%)} & \multicolumn{1}{c}{NFE} & \multicolumn{1}{c}{Rem.} & \multicolumn{1}{c}{Edit (\%)} \\
\midrule
\multirow{2}{*}{$2$} & Stride & $\bm{43.90}$ & $113$ & $51.6$ & $3.72$ & $\bm{77.40}$ & $110$ & $57.2$ & $3.18$ \\
& Random & $43.53{\scriptstyle\pm 0.6}$ \drop{0.37} & $119$ & $83.4$ & $3.30$ & $76.40{\scriptstyle\pm 0.9}$ \drop{1.00} & $115$ & $99.4$ & $2.74$ \\
\midrule
\multirow{2}{*}{$3$} & Stride & $43.29$ & $120$ & $37.6$ & $4.39$ & $76.80$ & $116$ & $38.5$ & $4.60$ \\
& Random & $43.29{\scriptstyle\pm 0.7}$ \neut{0.00} & $122$ & $53.7$ & $3.54$ & $76.80{\scriptstyle\pm 0.6}$ \neut{0.00} & $119$ & $59.6$ & $3.54$ \\
\midrule
\multirow{2}{*}{$4$} & Stride & $43.29$ & $127$ & $32.1$ & $4.95$ & $77.20$ & $123$ & $36.1$ & $4.54$ \\
& Random & $43.17{\scriptstyle\pm 0.2}$ \drop{0.12} & $129$ & $45.0$ & $4.02$ & $76.60{\scriptstyle\pm 0.6}$ \drop{0.60} & $125$ & $48.0$ & $3.90$ \\
\bottomrule
\end{tabular}
\end{table*}

\begin{figure*}[t]
\centering
\includegraphics[width=\textwidth]{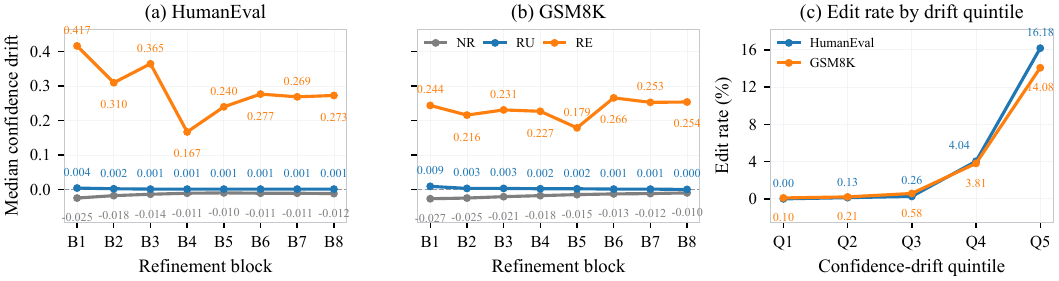}
\caption{Confidence-drift diagnostics. Panels (a)--(b) show median drift at each refinement boundary for tokens that are not remasked (NR), remasked but unchanged (RU), or remasked and edited (RE). Panel (c) reports the edit rate among positive-drift events grouped into equal-frequency drift quintiles.}
\label{fig:drift-diagnostics}
\end{figure*}

\section{Experiments}
\label{sec:experiments}
Our experiments are designed to answer five research questions:
\textbf{RQ1} Does CoDR improve performance across models, tasks, and base samplers?
\textbf{RQ2} Do the gains come from CoDR's core mechanisms, namely drift-guided triggering and structured partition probing?
\textbf{RQ3} What does confidence drift capture during decoding, and how is its magnitude related to token revision and successful correction?
\textbf{RQ4} How effectively and practically does CoDR use additional inference compute?
\textbf{RQ5} How robust is CoDR across hyperparameters and generation lengths?

\subsection{Experimental Setup}
\label{sec:setup-exp}

We evaluate two open-weight MDLMs, LLaDA-8B-Instruct and Dream-v0-Instruct-7B, on four zero-shot benchmarks: HumanEval \cite{chen2021evaluatinglargelanguagemodels}, MBPP \cite{austin2021programsynthesislargelanguage}, MATH-500 \cite{hendrycks2021measuring} , and GSM8K \cite{cobbe2021trainingverifierssolvemath}, use Pass@1 metric. For ablation study, we select a subset GSM8K with 500 samples. All runs use bf16 precision, generation length $L_{\mathrm{gen}}=256$, block length $32$, and 0-shot prompting. Unless otherwise stated, the main accuracy experiments do not use a KV cache; Table~\ref{tab:latency-kv} separately evaluates latency and memory with and without KV caching for the initial Fast-dLLM draft. Decoding uses argmax at temperature $0$ and is deterministic; reruns with different seeds produce identical predictions. All experiments are performed on NVIDIA A100 GPUs.

We evaluate Fast-dLLM (\texttt{fd}, $\tau=0.9$), EB-Sampler (\texttt{eb}, $\gamma=0.2$), and \texttt{top1} as standalone generation baselines. Unless noted, ``+CoDR'' uses Algorithm~\ref{alg:main} with $k=2$, $\tau_{\mathrm{drift}}=0$, parity split. In the refill stage, \texttt{fd+CoDR} uses Fast-dLLM with $\tau=0.9$, \texttt{eb+CoDR} uses EB-Sampler with $\gamma=0.1$, and \texttt{top1+CoDR} uses EB-Sampler with $\gamma=0.01$(a detailed analysis in Appendix \ref{sec:top1}). We additionally compare with ReMDM, following the setting used with LLaDA in the original paper, each block run $32$ ReMDM-loop steps, and CORE (detailed setting for CORE is shown in Appendix \ref{sec:core}). NFE (number of function evaluations) counts all decoding, probe, remasking, and refill forward passes.

\subsection{Main Results (RQ1)}
\label{sec:main-results}
Table~\ref{tab:main} shows that CoDR improves commitment-based decoding across both models and all three base samplers. On LLaDA-8B, adding CoDR improves the average accuracy of \texttt{top1}, \texttt{fd}, and \texttt{eb} by $+0.85$, $+1.06$, and $+1.44$ percentage points, respectively. On Dream-7B, the corresponding gains are $+0.82$, $+1.11$, and $+0.79$ percentage points. These consistent improvements indicate that CoDR is not specific to a particular decoding rule, but can be applied as a general refinement layer for different commitment-based samplers.

CoDR further improves the accuracy--compute trade-off. On LLaDA-8B, \texttt{fd+CoDR} achieves the best average accuracy of $48.93\%$ with $123$ NFE, outperforming ReMDM's $48.44\%$ average accuracy while using far fewer forward evaluations than its $512$ NFE setting. On Dream-7B, \texttt{top1+CoDR} obtains the best average accuracy of $60.62\%$ with $324$ NFE, and \texttt{fd+CoDR} reaches $60.20\%$ with only $135$ NFE. This suggests that CoDR uses additional computation in a targeted way by revisiting potentially unstable commitments, rather than uniformly increasing denoising iterations.

Because decoding is deterministic, uncertainty arises from the finite test sets rather than sampling variation. A $10{,}000$-resample paired bootstrap over LLaDA predictions finds significant four-task average gains for \texttt{top1}, Fast-dLLM, and EB-Sampler; all $95\%$ confidence intervals exclude zero. CoDR also improves both samplers on two additional benchmarks, BBH \cite{bbh} and GPQA \cite{rein2024gpqa}. Full tests and results are reported in Appendix ~\ref{sec:additional-results}.

\subsection{Effectiveness of Core Mechanisms (RQ2)}
\label{sec:rq2}

We examine whether CoDR's gains come from (i) confidence drift and (ii) structured dense-context probing.

\paragraph{Baselines.}
For triggering, we compare random remasking, \texttt{bottom-$m$} selection using only the lowest current confidence $\psi^{\mathrm{new}}$, \texttt{top1\_change}, and our temporal \texttt{conf\_drift} signal. The bottom-$m$ settings use $m=6$, $m=8$ and the random remasking use $p=0.045$, $p=0.20$ to approximately match CoDR's remasking budget. For probing, we compare deterministic stride and random partitions at the same $k$. We report accuracy, NFE, mean remasked positions (\textbf{Rem.}), and the percentage of remasked tokens that change after refill (\textbf{Edit}).

\paragraph{(i) Confidence Drift.}
Table~\ref{tab:random-trigger} shows that random remasking adds compute without improving accuracy, while \texttt{bottom-$m$} selection remains less accurate and more expensive than CoDR at comparable remasking budgets. The \texttt{top1\_change} trigger is precise but overly conservative, producing very few remasks. In contrast, \texttt{conf\_drift} achieves the best accuracy on both datasets, showing that temporal comparison provides information beyond current confidence alone.

\paragraph{(ii) Structured dense-context probing.}
Table~\ref{tab:partition-rand} shows that structured stride probing preserves more regular local context than random partitions. At $k{=}2$, random partitioning increases remasking from $51.6$ to $83.4$ on HumanEval and from $57.2$ to $99.4$ on GSM8K, while reducing accuracy from $43.90\%$ to $43.53\%$ and from $77.40\%$ to $76.40\%$. We therefore use deterministic stride probing by default.

\subsection{What Confidence Drift Captures (RQ3)}
\label{sec:rq3}

We analyze default CoDR with LLaDA-8B/Fast-dLLM on HumanEval and GSM8K ($L_{\mathrm{gen}}{=}256$, block size $32$, $k{=}2$, and $\tau_{\mathrm{drift}}{=}0$). A token-event is one committed position observed at one refinement boundary.

\paragraph{Dynamic evolution.}
At the end of each block B1--B8, CoDR probes every position committed at that boundary and computes $\delta_i=\psi_i^{\mathrm{old}}-\psi_i^{\mathrm{new}}$. We then classify each token-event by its outcome in that refinement round: \textbf{NR} was probed but not remasked (under $\tau_{\mathrm{drift}}=0$, typically $\delta_i\leq0$); \textbf{RU} had $\delta_i>0$ and was remasked, but refill reproduced the same token; and \textbf{RE} had $\delta_i>0$ and refill produced a different token. Figure~\ref{fig:drift-diagnostics}(a--b) reports the median $\delta$ of each group separately at every block. RE maintains substantially larger drift than the other groups from B1 to B8 on both tasks, showing that drift tracks evolving contextual instability rather than a calibrated error probability.

\paragraph{Drift magnitude and token revision.}
For each dataset, we collect all token-events with positive drift ($\delta>0$), which are selected for remasking under the default $\tau_{\mathrm{drift}}=0$. We sort these events by $\delta$ and divide them into five equal-frequency bins, from the lowest-drift quintile Q1 to the highest-drift quintile Q5. Within each bin, the edit rate is the fraction of events whose refill produces a token different from the original commitment. Figure~\ref{fig:drift-diagnostics}(c) shows that this rate rises sharply with drift, reaching $16.18\%$ on HumanEval and $14.08\%$ on GSM8K in Q5. Thus, drift magnitude identifies commitments that the model is more likely to overturn during refill. Full numerical results are reported in Appendix ~\ref{sec:drift-transition}.

\paragraph{Relation to successful correction.}
At the output level, Wrong$\rightarrow$Correct examples have the largest maximum and edited-token drift on GSM8K ($0.734/0.432$); HumanEval shows the same direction ($0.639/0.544$), although only two examples make this transition. Full correctness-transition statistics are reported in Appendix ~\ref{sec:drift-transition}.

\subsection{Practical Compute Efficiency (RQ4)}
\label{sec:rq4}

\begin{figure*}[t]
\centering
\begin{subfigure}[t]{0.45\textwidth}
  \centering
  \includegraphics[width=\linewidth]{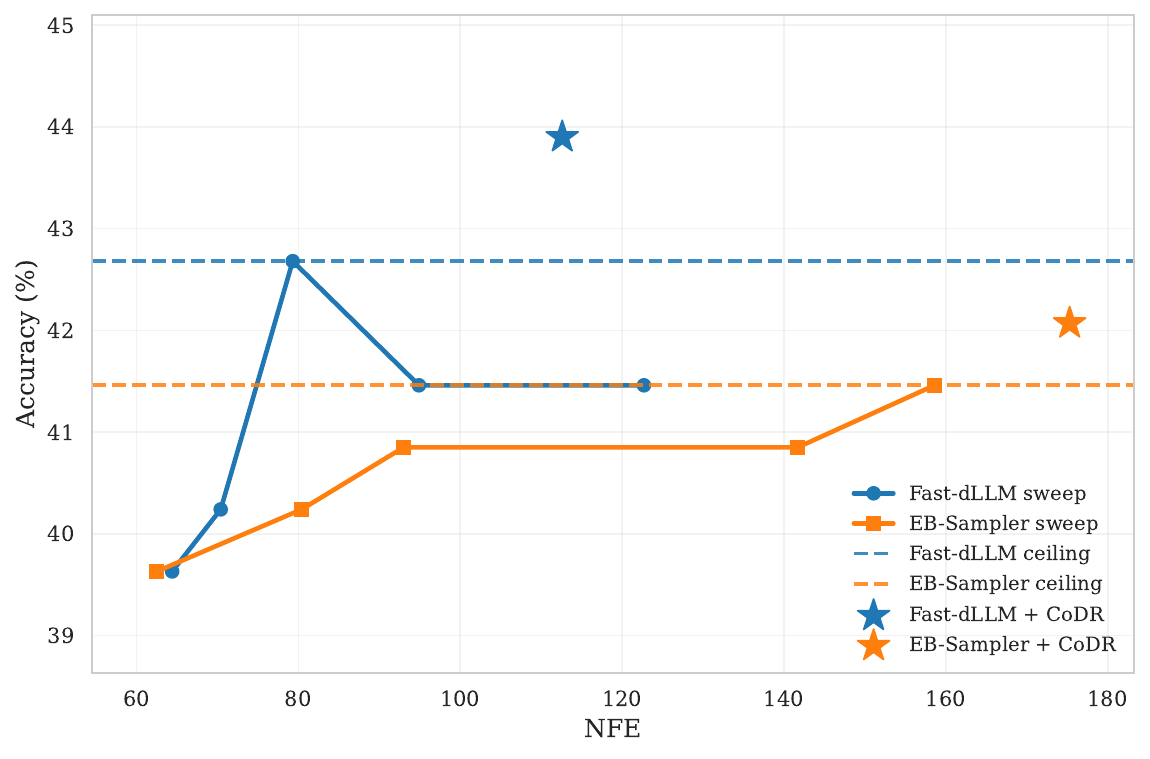}
  \caption{HumanEval}
  \label{fig:pareto_humaneval}
\end{subfigure}
\hfill
\begin{subfigure}[t]{0.45\textwidth}
  \centering
  \includegraphics[width=\linewidth]{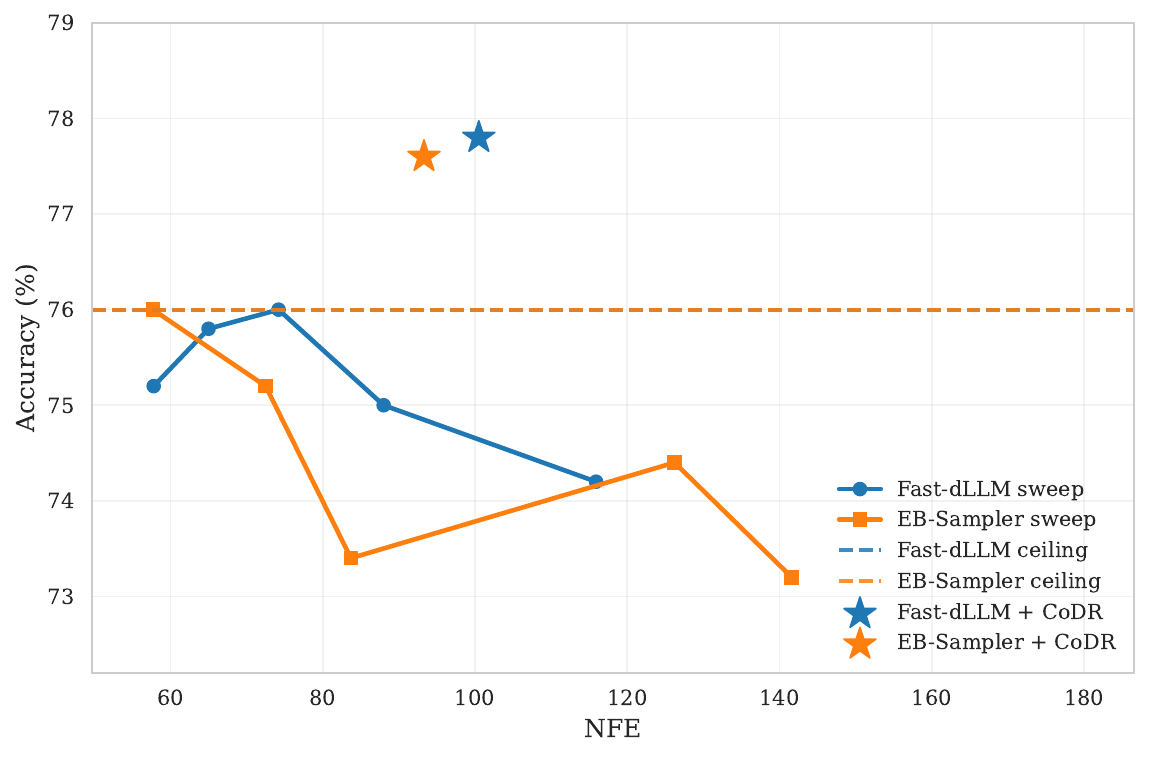}
  \caption{GSM8K}
  \label{fig:pareto_gsm8k}
\end{subfigure}
\caption{Compute allocation on LLaDA-8B ($0$-shot). Solid curves sweep the base sampler's compute knob, using $\tau$ for Fast-dLLM and $\gamma$ for EB-Sampler. Dashed lines mark the best swept baseline accuracy, and stars mark the best CoDR result. CoDR achieves higher accuracy than the swept baselines on both HumanEval and GSM8K, suggesting more effective use of additional NFE.}
\label{fig:pareto}
\end{figure*}

\paragraph{Compute allocation.}
Figure~\ref{fig:pareto} tests whether the base sampler can recover CoDR's gains by spending a similar NFE budget. Sweeping the Fast-dLLM threshold $\tau$ and EB-Sampler bound $\gamma$ reveals a clear scaling ceiling: additional base-sampler evaluations do not improve the best HumanEval or GSM8K accuracy. CoDR instead raises Fast-dLLM from $42.68\%$ to $43.90\%$ on HumanEval and from $76.00\%$ to $77.80\%$ on GSM8K, with the same trend for EB-Sampler. Thus, its gains come from targeted revision rather than additional NFE alone.

\begin{table}[t]
\centering
\small
\caption{Refinement schedules on LLaDA-8B/FD.}
\label{tab:refinement-schedule}
\begin{tabular*}{\columnwidth}{@{\extracolsep{\fill}}lrrrr@{}}
\toprule
& \multicolumn{2}{c}{HumanEval} & \multicolumn{2}{c}{GSM8K} \\
\cmidrule(lr){2-3}\cmidrule(lr){4-5}
Schedule & Acc. & NFE & Acc. & NFE \\
\midrule
None & $42.68$ & $79.0$ & $76.00$ & $74.0$ \\
Every 1 & $43.90$ & $112.3$ & $77.40$ & $110.0$ \\
Every 2 & $\bm{43.90}$ & $\bm{97.1}$ & $\bm{77.60}$ & $\bm{93.2}$ \\
Every 4 & $43.29$ & $89.2$ & $77.00$ & $84.4$ \\
Final & $43.29$ & $85.3$ & $76.20$ & $79.7$ \\
Mid-block & $42.68$ & $142.3$ & $76.40$ & $141.8$ \\
\bottomrule
\end{tabular*}
\end{table}

\paragraph{Refinement schedule.}
Table~\ref{tab:refinement-schedule} shows that every-block refinement is not required. Refining every two blocks preserves HumanEval accuracy, slightly improves GSM8K, and reduces NFE by $14$--$15\%$. Less frequent schedules trade accuracy for further savings, whereas mid-block refinement exceeds $140$ NFE without a meaningful gain. We retain every block as the simple default and recommend every two blocks when efficiency matters.

\begin{table}[t]
\centering
\small
\setlength{\tabcolsep}{2pt}
\caption{HumanEval latency and memory on LLaDA-8B/FD ($L_{\mathrm{gen}}{=}256$, batch size $1$). E$n$: refine every $n$ blocks; F: final only.}
\label{tab:latency-kv}
\begin{tabular*}{\columnwidth}{@{\extracolsep{\fill}}llrrrr@{}}
\toprule
Cache & Method & Acc. & NFE & Sec. & MiB \\
\midrule
\multirow{5}{*}{No KV}
& FD & $42.68$ & $79.09$ & $4.40$ & $17{,}512$ \\
& CoDR-E1 & $43.90$ & $112.27$ & $6.36$ & $17{,}512$ \\
& CoDR-E2 & $\bm{43.90}$ & $97.02$ & $\bm{5.55}$ & $17{,}512$ \\
& CoDR-E4 & $43.29$ & $89.23$ & $5.10$ & $17{,}512$ \\
& CoDR-F & $43.29$ & $85.32$ & $4.88$ & $17{,}512$ \\
\midrule
\multirow{5}{*}{KV draft}
& FD & $43.90$ & $82.82$ & $2.77$ & $18{,}322$ \\
& CoDR-E1 & $45.12$ & $119.02$ & $4.79$ & $18{,}222$ \\
& CoDR-E2 & $\bm{45.12}$ & $102.20$ & $\bm{3.90}$ & $18{,}222$ \\
& CoDR-E4 & $43.29$ & $94.41$ & $3.94$ & $18{,}182$ \\
& CoDR-F & $43.90$ & $89.01$ & $3.18$ & $18{,}182$ \\
\bottomrule
\end{tabular*}
\end{table}

\paragraph{Latency, memory, and KV caching.}
Table~\ref{tab:latency-kv} confirms that NFE reflects wall-clock cost. Every-two-block CoDR improves accuracy by $1.22$ points while increasing latency from $4.40$ to $5.55$ seconds without KV caching, and from $2.77$ to $3.90$ seconds with KV drafting. This schedule is faster than every-block refinement in both settings and does not increase peak memory. KV drafting currently accelerates only initial sampling because remasking invalidates cached states for probing and refill.

\begin{figure*}[t]
\centering
\includegraphics[width=\textwidth]{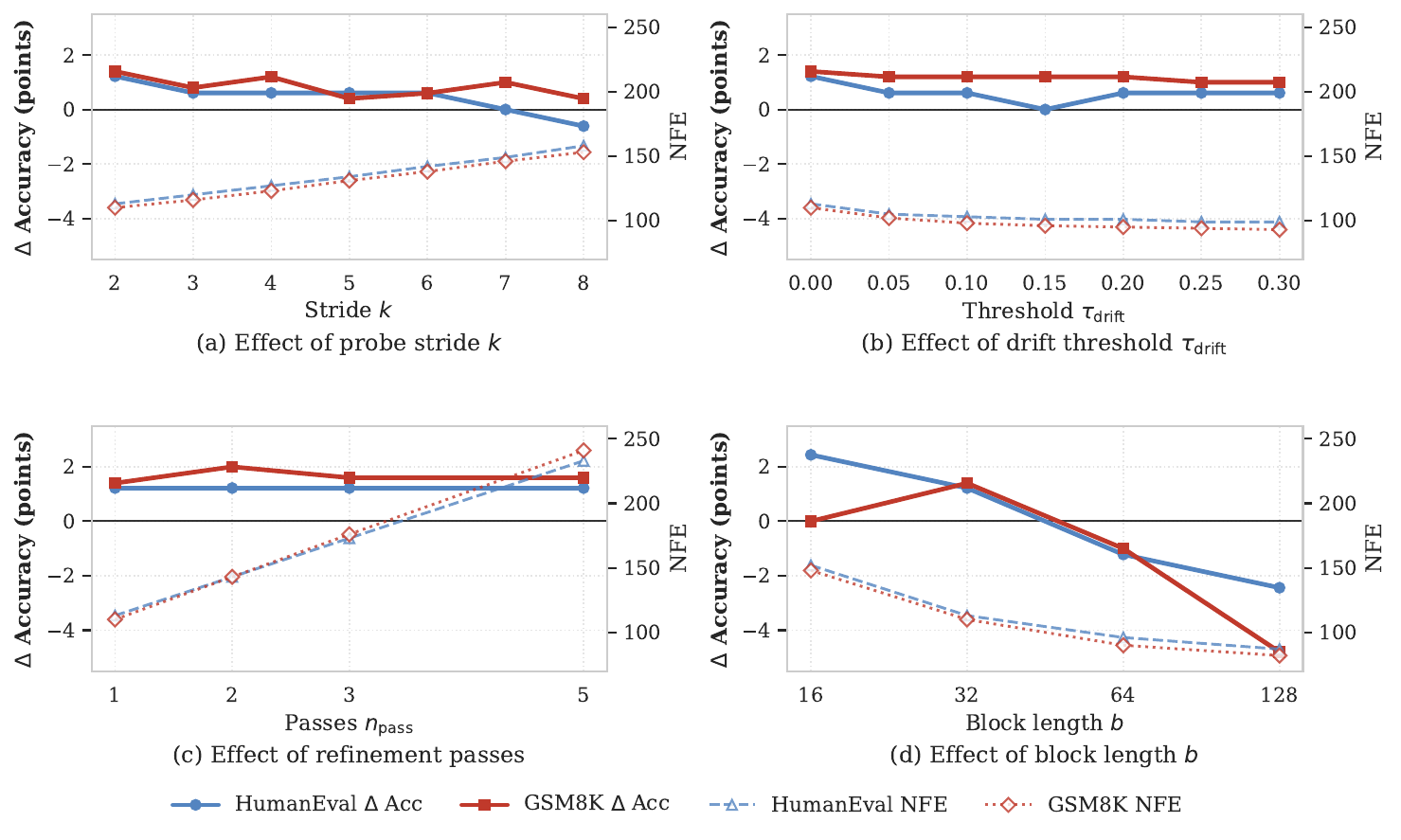}
\caption{ablation study for our method.}
\vspace{-10pt}
\label{fig:ablation}
\end{figure*}

\subsection{Ablation Study (RQ5)}
\label{sec:ablation}
\paragraph{Effect of probe stride $k$.}
The stride $k$ controls the number of partition groups used for probing. Larger $k$ preserves denser context in each probe but requires more forward evaluations. Figure~\ref{fig:ablation}(a), with detailed results in Table~\ref{tab:ablation_stride_k}, shows that $k=2$ gives the best accuracy--NFE trade-off, while larger values add compute without clear accuracy gains. We therefore use $k=2$ as the default.

\paragraph{Effect of drift threshold.}
The drift threshold $\tau_{\mathrm{drift}}$ controls how conservative the remasking trigger is. The default $\tau_{\mathrm{drift}}=0$ remasks any token whose confidence drops under the probe context, while positive thresholds filter out small-drift tokens. Figure~\ref{fig:ablation}(b), with detailed results in Table~\ref{tab:ablation_tau}, shows that $\tau_{\mathrm{drift}}=0$ achieves the best accuracy on both HumanEval and GSM8K. Positive thresholds reduce NFE but do not improve accuracy, so we keep $\tau_{\mathrm{drift}}=0$ as the default.

\paragraph{Effect of refinement passes.}
We ablate the number of remask--refill passes at each block boundary. Figure~\ref{fig:ablation}(c), with detailed results in Table~\ref{tab:ablation_npass}, shows that one pass is generally sufficient. Additional passes substantially increase NFE; they leave HumanEval accuracy unchanged and bring only a small gain on GSM8K with two passes. This suggests that the first pass captures most useful drift signals, so we use one refinement pass by default.

\paragraph{Effect of block length.}
The block length $b$ determines how often refinement is invoked. Smaller blocks create more refinement opportunities, while larger blocks reduce intervention frequency and NFE. Figure~\ref{fig:ablation}(d), with detailed results in Table~\ref{tab:block_ablation}, shows that larger blocks degrade the base sampler, suggesting that sparse refinement boundaries allow early errors to accumulate. CoDR improves over the corresponding baseline across most block lengths. We use $b=32$ as the default because it gives the best GSM8K result and a strong HumanEval result at lower NFE than $b=16$, consistent with recent work~\cite{wu2025fastdllm}.

\begin{figure}[t]
\centering
\includegraphics[width=\columnwidth]{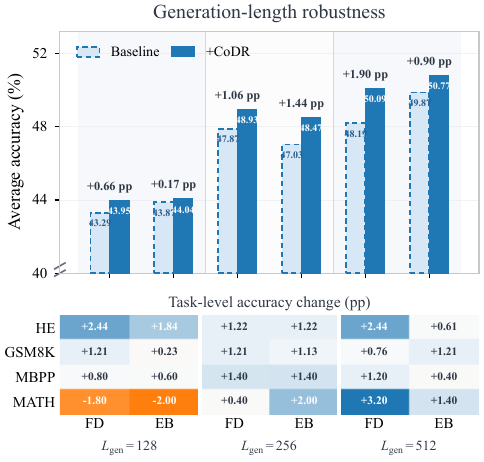}
\caption{\textbf{Generation-length robustness on LLaDA-8B.}
Four-task average accuracy of the baseline and CoDR under Fast-dLLM (FD) and EB sampling across generation lengths. The heatmap reports task-level accuracy gains (pp) from adding CoDR.}
\label{fig:generation-length}
\end{figure}

\paragraph{Robustness across generation lengths.}
Figure~\ref{fig:generation-length} compares lengths $128$, $256$, and $512$ across all four tasks. CoDR improves average accuracy for both samplers at every length, with gains of $1.90$ points for Fast-dLLM and $0.90$ for EB-Sampler at length $512$. At length $128$, gains are smaller and MATH-500 regresses, showing robust average improvements but not uniform per-task gains. NFE overhead increases with sequence length. Appendix Table~\ref{tab:generation-length} provides full results.

\section{Conclusion}
\label{sec:conclusion}
We introduce CoDR, a training-free method that revisits MDLM tokens whose confidence decreases as context grows. Its $k$-partition probe checks all commitments in a few forward passes, then selectively refills unstable tokens under richer context. Across two backbones, three samplers, and math and code tasks, CoDR improves the accuracy--compute trade-off. Diagnostics associate larger drift with token edits and successful corrections, while controlled comparisons distinguish targeted revision from extra decoding alone. These results show that reversible decoding can correct early commitments without retraining or replacing the base sampler. Future work may explore adaptive probing and cache-aware refinement.

\clearpage

\section*{Limitations}

CoDR is an inference-time refinement method whose effectiveness depends on the underlying base sampler used for refill. The confidence trigger identifies remasked positions whose support decreases under denser context, but the quality of revised tokens remains bounded by the same MDLM and sampler used for generation. As a result, occasional regressions may occur when refill introduces fresh local errors, especially in arithmetic-heavy cases. In addition, confidence drift is designed as a targeted revision signal rather than a complete error detector. CoDR intentionally revises only tokens whose support decreases under denser context, leaving other tokens unchanged to avoid excessive editing. Future work could explore adaptive refinement schedules, stronger refill constraints, and broader evaluation on larger models, multilingual tasks, and KV-cache optimized decoding.

\section*{Acknowledgments}

The authors acknowledge support from The Hong Kong Polytechnic University (Research Grant P0046811), the National Natural Science Foundation of China under Grant No. 62136005, and the Shenzhen Fundamental Research Program under Grant No. JCYJ20250604144724032.


\bibliography{codr}

\clearpage

\appendix

\section{Detailed Experiment Setting}
Unless otherwise noted, all experiments follow the model, decoding, precision, prompting, and hardware settings described in Section~\ref{sec:setup-exp}. Sampler-specific implementation details are provided in the later appendix sections.

\subsection{Model and Dataset Licenses}
All pre-trained models (LLaDA and Dream) and datasets (HumanEval, MBPP, GSM8K, and MATH-500) used in this work are publicly available under open-source licenses, such as MIT or Apache 2.0, and are used strictly for academic research and methodological evaluation, consistent with their intended use.

\section{Statistical Reliability and Additional Benchmarks}
\label{sec:additional-results}

\subsection{Paired Bootstrap Analysis}
Our argmax decoding at temperature $0$ is deterministic: rerunning each setting with different seeds produces identical predictions and accuracies. We therefore quantify finite-test-set uncertainty using paired bootstrap resampling. For each LLaDA sampler, we resample paired baseline/CoDR predictions within each task $10{,}000$ times, compute the accuracy gain per task, and average the four gains equally to match Table~\ref{tab:main}. Two-sided results are reported in percentage points in Table~\ref{tab:bootstrap}. All confidence intervals exclude zero, indicating a significant average gain for each sampler.

\begin{table}[H]
\centering
\scriptsize
\setlength{\tabcolsep}{1.5pt}
\caption{Paired bootstrap tests for the LLaDA four-task average ($10{,}000$ resamples).}
\label{tab:bootstrap}
\begin{tabular*}{\columnwidth}{@{\extracolsep{\fill}}lrrrrr@{}}
\toprule
Sampler & Base & CoDR & $\Delta$ pp & $95\%$ CI & $p$ \\
\midrule
\texttt{top1} & $47.84$ & $48.69$ & $+0.85$ & $[0.12,1.65]$ & $0.026$ \\
Fast-dLLM & $47.87$ & $48.93$ & $+1.06$ & $[0.33,1.84]$ & $0.006$ \\
EB-Sampler & $47.03$ & $48.47$ & $+1.44$ & $[0.63,2.33]$ & $<0.001$ \\
\bottomrule
\end{tabular*}
\end{table}

\subsection{Additional Benchmarks}
We further evaluate LLaDA-8B on BBH and GPQA. As shown in Table~\ref{tab:additional-benchmarks}, CoDR improves accuracy over both Fast-dLLM and EB-Sampler on both benchmarks, extending the main results beyond the original four tasks.

\begin{table}[H]
\centering
\small
\caption{Additional benchmark results. Each cell reports accuracy (\%)/NFE.}
\label{tab:additional-benchmarks}
\begin{tabular*}{\columnwidth}{@{\extracolsep{\fill}}lcc@{}}
\toprule
Method & BBH & GPQA \\
\midrule
FD & $50.94/64.48$ & $21.72/138.45$ \\
CoDR-FD & $\bm{51.51}/87.48$ & $\bm{22.73}/172.90$ \\
EB & $51.51/79.31$ & $21.21/142.49$ \\
CoDR-EB & $\bm{51.73}/95.18$ & $\bm{22.22}/176.40$ \\
\bottomrule
\end{tabular*}
\end{table}

\FloatBarrier

\section{Confidence Drift and Correctness Transitions}
\label{sec:drift-transition}

\subsection{Token-Level Drift Diagnostics}
Tables~\ref{tab:drift-trajectory} and~\ref{tab:drift-quintiles} provide the numerical results underlying Figure~\ref{fig:drift-diagnostics}: per-block median drift by refinement outcome and edit rates across positive-drift quintiles, respectively.

\newcommand{\DriftTrajectoryTable}{%
\begin{table}[H]
\centering
\scriptsize
\caption{Per-block median confidence drift by refinement outcome. NR: not remasked; RU: remasked but unchanged; RE: remasked and edited.}
\label{tab:drift-trajectory}
\setlength{\tabcolsep}{0.8pt}
\resizebox{\columnwidth}{!}{%
\begin{tabular}{lrrrrrrrr}
\toprule
\multicolumn{9}{c}{HumanEval} \\
\midrule
Group & B1 & B2 & B3 & B4 & B5 & B6 & B7 & B8 \\
\midrule
NR & $-0.025$ & $-0.018$ & $-0.014$ & $-0.011$ & $-0.010$ & $-0.011$ & $-0.011$ & $-0.012$ \\
RU & $0.004$ & $0.002$ & $0.001$ & $0.001$ & $0.001$ & $0.001$ & $0.001$ & $0.001$ \\
\textbf{RE} & $\bm{0.417}$ & $\bm{0.310}$ & $\bm{0.365}$ & $\bm{0.167}$ & $\bm{0.240}$ & $\bm{0.277}$ & $\bm{0.269}$ & $\bm{0.273}$ \\
\midrule
\multicolumn{9}{c}{GSM8K} \\
\midrule
Group & B1 & B2 & B3 & B4 & B5 & B6 & B7 & B8 \\
\midrule
NR & $-0.027$ & $-0.025$ & $-0.021$ & $-0.018$ & $-0.015$ & $-0.013$ & $-0.012$ & $-0.010$ \\
RU & $0.009$ & $0.003$ & $0.003$ & $0.002$ & $0.002$ & $0.001$ & $0.001$ & $0.000$ \\
\textbf{RE} & $\bm{0.244}$ & $\bm{0.216}$ & $\bm{0.231}$ & $\bm{0.227}$ & $\bm{0.179}$ & $\bm{0.266}$ & $\bm{0.253}$ & $\bm{0.254}$ \\
\bottomrule
\end{tabular}%
}
\end{table}
}

\newcommand{\DriftQuintilesTable}{%
\begin{table}[H]
\centering
\small
\setlength{\tabcolsep}{2pt}
\caption{Edit rate (\%) among positive-drift events grouped by equal-frequency drift quintile.}
\label{tab:drift-quintiles}
\begin{tabular*}{\columnwidth}{@{\extracolsep{\fill}}lrrrrr@{}}
\toprule
Data & Q1 & Q2 & Q3 & Q4 & Q5 \\
\midrule
HumanEval & $0$ & $0.13$ & $0.26$ & $4.04$ & $\bm{16.18}$ \\
GSM8K & $0.10$ & $0.21$ & $0.58$ & $3.81$ & $\bm{14.08}$ \\
\bottomrule
\end{tabular*}
\end{table}
}

\subsection{Output-Level Correctness Transitions}

We pair each initial Fast-dLLM output with its final CoDR output and group examples by correctness transition. For each example, Max $\delta$ is the largest token-event drift observed during refinement, and Edited-token $\delta$ is the median drift over events whose refill changes the token. Examples without edited events are excluded from the latter statistic. Table~\ref{tab:correctness-transition} reports the group median for each measure.

\newcommand{\CorrectnessTransitionTable}{%
\begin{table}[H]
\centering
\scriptsize
\setlength{\tabcolsep}{1.5pt}
\caption{Confidence drift by output correctness transition.}
\label{tab:correctness-transition}
\begin{tabular*}{\columnwidth}{@{\extracolsep{\fill}}llrrr@{}}
\toprule
Data & Transition & $n$ & Max $\delta$ & Edited-token $\delta$ \\
\midrule
\multirow{4}{*}{GSM8K}
& Wrong$\rightarrow$Correct & $13$ & $\bm{0.734}$ & $\bm{0.432}$ \\
& Correct$\rightarrow$Correct & $374$ & $0.404$ & $0.282$ \\
& Wrong$\rightarrow$Wrong & $107$ & $0.527$ & $0.287$ \\
& Correct$\rightarrow$Wrong & $6$ & $0.627$ & $0.376$ \\
\midrule
\multirow{4}{*}{HumanEval}
& Wrong$\rightarrow$Correct & $2$ & $\bm{0.639}$ & $\bm{0.544}$ \\
& Correct$\rightarrow$Correct & $70$ & $0.461$ & $0.314$ \\
& Wrong$\rightarrow$Wrong & $92$ & $0.494$ & $0.311$ \\
& Correct$\rightarrow$Wrong & $0$ & --- & --- \\
\bottomrule
\end{tabular*}
\end{table}
}

On GSM8K, Wrong$\rightarrow$Correct examples exhibit higher maximum and edited-token drift than the other transition groups. HumanEval follows the same direction, although the Wrong$\rightarrow$Correct group contains only two examples. Together with Table~\ref{tab:drift-quintiles}, these results associate larger drift with both actual token revision and successful output correction.

\subsection{Top1 Setting and Analysis}
\label{sec:top1}
Both variants of our Top1 refinement procedure use top-$1$ decoding for the initial
generation pass; they differ only in how remasked positions are \emph{refilled} during revision. We compare refilling with top-$1$ decoding against refilling with an entropy-bounded (EB) sampler, holding everything else fixed (top-$1$ phase-1 generation, identical probe and remask schedule). Top-$1$ refill commits exactly one token per forward pass, so every remasked span is regenerated one token at a time, adding one forward pass per remasked token and accumulating large number of function evaluations (NFE) across all blocks and revision rounds. The EB sampler can refill multiple tokens in one forward pass and therefore requires far fewer passes.

This change is free in terms of quality. With the phase-1 top-$1$ generation held
fixed, top-$1$ and EB refill produce \emph{identical} accuracy---pass@1 $0.4268$
on LLaDA-8B HumanEval and $0.5915$ on Dream-7B HumanEval---while
EB refill reduces NFE from $377$ to $303$ on LLaDA and from $393$ to $317$ on
Dream, a $\sim\!20\%$ saving in both cases (Table~\ref{tab:refill}). We therefore
use top-$1$ generation with EB refill ($\gamma=0.01$) as our default: it keeps the
high-precision, one-token-at-a-time ordering of top-$1$ where decision quality
matters most---the main generation pass---while avoiding its redundant cost when
re-filling remasked spans.

\newcommand{\RefillComparisonTable}{%
\begin{table}[H]
\centering
\small
\caption{Refill sampler comparison with top-$1$ generation held fixed
(HumanEval, 0-shot). Top-$1$ and EB refill give identical accuracy; EB refill
cuts NFE by $\sim\!20\%$.}
\label{tab:refill}
\begin{tabular*}{\columnwidth}{@{\extracolsep{\fill}}llcc@{}}
\toprule
Model & Refill & pass@1 & NFE \\
\midrule
\multirow{2}{*}{LLaDA-8B} & top-$1$ & 0.4268 & 377 \\
                          & EB      & 0.4268 & \textbf{303} \\
\midrule
\multirow{2}{*}{Dream-7B} & top-$1$ & 0.5915 & 393 \\
                          & EB      & 0.5915 & \textbf{317} \\
\bottomrule
\end{tabular*}
\end{table}
}

\subsection{CoRE Setting and Analysis}
\label{sec:core}
We follow the CoRE protocol~\cite{zhai2026core}: generation length $256$,
top-$1$ decoding (matching the baseline), revision every $E=8$ steps within
the mid-trajectory window $[0.25, 0.75]$, $k=1$ token revised per step, and a
replacement-confidence threshold of $0.3$. Unlike the original paper, which
decodes the entire response as a \emph{single} block, we adopt \emph{block-wise}
decoding (block $=32$) with candidate-set size $m=4$ and $\mathrm{protect}=8$
for \emph{both} LLaDA and Dream.

\paragraph{Block-wise decoding outperforms single-block on LLaDA.}
While the paper's single-block configuration runs without collapse on LLaDA, it
is overall inferior to block-wise decoding, particularly on multi-step
reasoning. Table~\ref{tab:core-block} compares the two at matched compute
(NFE $\approx 272$). Single-block is marginally stronger on HumanEval, where the
larger global context aids code structure, but block-wise decoding is
substantially better on MBPP, MATH-500, and GSM8K, raising the four-task average
from $0.394$ to $0.469$. The gap is most pronounced on reasoning: single-block
with $m=32$ degrades MATH-500 to $0.248$ and GSM8K to $0.536$, whereas
block-wise decoding with $m=4$ recovers them to $0.374$ and $0.723$, near the
no-revision baseline. We therefore use block-wise decoding for LLaDA as well.

\newcommand{\CoreBlockTable}{%
\begin{table}[H]
\centering
\small
\caption{LLaDA-8B-Instruct CoRE: single-block ($m{=}32$) vs.\ block-wise
($m{=}4$), 0-shot, NFE $\approx 272$. Block-wise decoding wins on three of four
tasks and on average, with the largest gains on reasoning.}
\label{tab:core-block}
\setlength{\tabcolsep}{1.5pt}
\begin{tabular*}{\columnwidth}{@{\extracolsep{\fill}}lccccc@{}}
\toprule
Setting & HE & MBPP & MATH & GSM8K & Avg. \\
\midrule
Single block, $m{=}32$ & \textbf{0.427} & 0.364 & 0.248 & 0.536 & 0.394 \\
Block-wise, $m{=}4$    & 0.384 & \textbf{0.394} & \textbf{0.374} & \textbf{0.723} & \textbf{0.469} \\
\bottomrule
\end{tabular*}
\end{table}
}

\paragraph{Single-block decoding fails outright on Dream.}
On Dream the single-block configuration does not merely underperform---it
collapses to near-empty generations (with accuracy only $0.189$ in humaneval). This is not a CoRE-specific implementation
artifact: Dream's own validated samplers (\texttt{fastdllm}, \texttt{top1}) also
collapse under single-block decoding ($\sim\!12$ output tokens). The failure is a
property of Dream under single-block decoding---given a single fully-masked
block, the model places high confidence on an early end-of-sequence token before
any context has been committed, terminating generation prematurely. Block-wise
decoding (block $=32$) removes this collapse.

\paragraph{Candidate-set size must scale with block length.}
Within block-wise decoding, the paper's $m=32$ is far too large: masking $32$
candidates inside a $32$-token block erases almost the entire committed context
at revision time. The verify pass then scores token stability under a near-empty
context, mis-flags already-committed tokens, and derails generation (on Dream
HumanEval, pass@1 $0.5732 \rightarrow 0.3598$; the same over-perturbation drives
LLaDA's single-block reasoning collapse above). We therefore rescale $m$ by
\emph{perturbation ratio} rather than absolute count: the paper's $m=32$ perturbs
roughly $25\%$ of the $\sim\!128$ tokens committed at mid-window revision time,
which on a $32$-token block ($\sim\!16$ tokens filled at that point) corresponds
to $m \approx 4$. With block $=32$ and $m=4$, both models produce normal-length
outputs and recover baseline-level accuracy.

\DriftTrajectoryTable
\DriftQuintilesTable
\CorrectnessTransitionTable
\RefillComparisonTable
\CoreBlockTable
\FloatBarrier

\FloatBarrier
\section{Comprehensive Ablation Results}
\label{sec:ablation_table}

This section provides the exact numerical data corresponding to the hyperparameter sensitivity analysis visualized in Section~\ref{sec:ablation}. While the main text focuses on the overall Pareto-optimal trends and performance trade-offs, the tables below detail the discrete Accuracy and NFE metrics across all evaluated configurations of the CoDR framework.

Table~\ref{tab:generation-length} reports robustness across generation lengths. Table~\ref{tab:ablation_stride_k} reports the isolated effects of probe stride $k$, drift threshold $\tau_{\mathrm{drift}}$(Table \ref{tab:ablation_tau}), and the number of refinement passes $n_{\mathrm{pass}}$ (Table \ref{tab:ablation_npass}). Table~\ref{tab:block_ablation} details the impact of block length $b$, explicitly comparing the baseline generation against the refined output to illustrate the compounding effect of early errors in sparse commit boundaries. All results are evaluated using the Fast-dLLM base sampler (\texttt{fd}, $\tau{=}0.9$) on LLaDA-8B under a $0$-shot setting.

\begin{table}[H]
\centering
\scriptsize
\setlength{\tabcolsep}{1.2pt}
\caption{Full numerical results for generation-length robustness on LLaDA-8B. Task columns report accuracy (\%); averages are unweighted over HumanEval (HE), full GSM8K, MBPP, and MATH-500.}
\label{tab:generation-length}
\begin{tabular*}{\columnwidth}{@{\extracolsep{\fill}}lrrrrrr@{}}
\toprule
\multicolumn{7}{c}{$L_{\mathrm{gen}}=128$} \\
\midrule
Method & HE & GSM8K & MBPP & MATH & Avg. Acc. & Avg. NFE \\
\midrule
FD & $33.54$ & $69.22$ & $36.60$ & $33.80$ & $43.29$ & $47.50$ \\
$+$CoDR & $35.98$ & $70.43$ & $37.40$ & $32.00$ & $\bm{43.95}$ & $64.96$ \\
EB & $34.14$ & $69.75$ & $38.00$ & $33.60$ & $43.87$ & $55.78$ \\
$+$CoDR & $35.98$ & $69.98$ & $38.60$ & $31.60$ & $\bm{44.04}$ & $65.88$ \\
\midrule
\multicolumn{7}{c}{$L_{\mathrm{gen}}=512$} \\
\midrule
FD & $42.68$ & $75.89$ & $38.00$ & $36.20$ & $48.19$ & $144.29$ \\
$+$CoDR & $45.12$ & $76.65$ & $39.20$ & $39.40$ & $\bm{50.09}$ & $227.48$ \\
EB & $45.73$ & $75.74$ & $39.20$ & $38.80$ & $49.87$ & $143.25$ \\
$+$CoDR & $46.34$ & $76.95$ & $39.60$ & $40.20$ & $\bm{50.77}$ & $234.48$ \\
\bottomrule
\end{tabular*}
\end{table}

\begin{table}[H]
\centering
\small
\caption{Ablation results on probe stride $k$ using \texttt{fd\_refine} with LLaDA-8B. Accuracy is reported in percentage. Bold values indicate optimal configurations.}
\label{tab:ablation_stride_k}
\begin{tabular*}{\columnwidth}{@{\extracolsep{\fill}}r r r r r@{}}
\toprule
& \multicolumn{2}{c}{HumanEval} & \multicolumn{2}{c}{GSM8K} \\
\cmidrule(lr){2-3} \cmidrule(lr){4-5}
$k$ & Acc (\%) & NFE & Acc (\%) & NFE \\
\midrule
$2$ & $\bm{43.90}$ & $113$ & $\bm{77.40}$ & $110$ \\
$3$ & $43.29$ & $120$ & $76.80$ & $116$ \\
$4$ & $43.29$ & $127$ & $77.20$ & $123$ \\
$5$ & $43.29$ & $134$ & $76.40$ & $131$ \\
$6$ & $43.29$ & $142$ & $76.60$ & $138$ \\
$7$ & $42.68$ & $149$ & $77.00$ & $146$ \\
$8$ & $42.07$ & $158$ & $76.40$ & $153$ \\
\bottomrule
\end{tabular*}
\end{table}

\begin{table}[H]
\centering

\small
\caption{Ablation results on drift threshold $\tau_{\mathrm{drift}}$ using \texttt{fd\_refine} with LLaDA-8B. Positive thresholds reduce NFE by making the trigger more selective.}
\label{tab:ablation_tau}
\begin{tabular*}{\columnwidth}{@{\extracolsep{\fill}}r r r r r@{}}
\toprule
& \multicolumn{2}{c}{HumanEval} & \multicolumn{2}{c}{GSM8K} \\
\cmidrule(lr){2-3} \cmidrule(lr){4-5}
$\tau_{\mathrm{drift}}$ & Acc (\%) & NFE & Acc (\%) & NFE \\
\midrule
$0.00$  & $\bm{43.90}$ & $113$ & $\bm{77.40}$ & $110$ \\
$0.05$  & $43.29$ & $105$ & $77.20$ & $102$ \\
$0.10$  & $43.29$ & $103$ & $77.20$ & $\phantom{0}98$ \\
$0.15$  & $42.68$ & $101$ & $77.20$ & $\phantom{0}96$ \\
$0.20$  & $43.29$ & $101$ & $77.20$ & $\phantom{0}95$ \\
$0.25$  & $43.29$ & $\phantom{0}99$ & $77.00$ & $\phantom{0}94$ \\
$0.30$  & $43.29$ & $\phantom{0}99$ & $77.00$ & $\phantom{0}93$ \\
\bottomrule
\end{tabular*}
\end{table}

\begin{table}[H]
\centering
\small
\caption{Ablation results on refinement passes $n_{\mathrm{pass}}$ using \texttt{fd\_refine} with LLaDA-8B. Additional refinement passes beyond $n=1$ add substantial compute for little or no gain.}
\label{tab:ablation_npass}
\begin{tabular*}{\columnwidth}{@{\extracolsep{\fill}}r r r r r@{}}
\toprule
& \multicolumn{2}{c}{HumanEval} & \multicolumn{2}{c}{GSM8K} \\
\cmidrule(lr){2-3} \cmidrule(lr){4-5}
$n_{\mathrm{pass}}$ & Acc (\%) & NFE & Acc (\%) & NFE \\
\midrule
$1$ & $\bm{43.90}$ & $113$ & $77.40$ & $110$ \\
$2$ & $43.90$ & $143$ & $\bm{78.00}$ & $143$ \\
$3$ & $43.90$ & $173$ & $77.60$ & $176$ \\
$5$ & $43.90$ & $233$ & $77.60$ & $241$ \\
\bottomrule
\end{tabular*}
\end{table}

\begin{table}[H]
\centering
\small
\caption{Ablation on block length $b$ using \texttt{fd\_refine} with LLaDA-8B. Block length controls the intervention frequency and the accuracy--NFE trade-off. Extreme sparse boundaries ($b=128$) allow early contextual errors to compound, degrading baseline performance and limiting refinement capabilities.}
\label{tab:block_ablation}
\begin{tabular*}{\columnwidth}{@{\extracolsep{\fill}}lrrrr@{}}
\toprule
\multicolumn{5}{c}{HumanEval} \\
\midrule
$b$ & Base A. & Base N. & CoDR A. & CoDR N. \\
\midrule
$16$ & $\bm{44.51}$ & $87$ & $\bm{45.12}$ & $152$ \\
$32$ \textbf{(def.)} & $42.68$ & $79$ & $43.90$ & $113$ \\
$64$ & $41.46$ & $77$ & $41.46$ & $96$ \\
$128$ & $39.02$ & $76$ & $40.24$ & $87$ \\
\midrule
\multicolumn{5}{c}{GSM8K} \\
\midrule
$b$ & Base A. & Base N. & CoDR A. & CoDR N. \\
\midrule
$16$ & $75.20$ & $81$ & $76.00$ & $148$ \\
$32$ \textbf{(def.)} & $76.00$ & $74$ & $\bm{77.40}$ & $110$ \\
$64$ & $73.20$ & $71$ & $75.00$ & $90$ \\
$128$ & $69.60$ & $72$ & $71.20$ & $82$ \\
\bottomrule
\end{tabular*}
\end{table}

\section{Detailed Case Studies}
\label{sec:appendix_cases}

To provide a comprehensive analysis of our method, this section presents detailed generation trajectories across mathematics and code generation tasks. We contrast the base Fast-dLLM sampler ($\tau=0.9$) against our CoDR formulation on LLaDA-8B in setting.

\subsection{Success Cases}
CoDR excels at identifying and correcting early irrational commitments induced by sparse context, cleanly overwriting them once the bidirectional global context stabilizes.

\paragraph{Code Generation Recovery.}
Figures~\ref{fig:success_code_1} and \ref{fig:success_code_2} illustrate CoDR's capacity to repair localized syntactic and API-level fractures. In Figure~\ref{fig:success_code_1} (HumanEval/89), the base sampler commits to a malformed character shift logic due to premature evaluation. CoDR perfectly isolates the confidence drift precisely at this arithmetic expression and recovers the exact \texttt{+4} rotation algorithm. Similarly, Figure~\ref{fig:success_code_2} demonstrates the recovery of a missing dependency; CoDR detects the structural vulnerability of invoking \texttt{math.ceil} without an import and seamlessly prepends \texttt{import math} during the refinement pass.

\paragraph{Mathematical Reasoning Correction.}
Figures~\ref{fig:success_math_1} and \ref{fig:success_math_2} highlight CoDR's precision in arithmetic and algebraic contexts. In Figure~\ref{fig:success_math_1}, a pure arithmetic slip ($1764/12=142$) derails the entire downstream revenue projection. CoDR flags the fractional setup, unmasks the flawed quotient, and enforces the mathematically sound answer. More impressively, Figure~\ref{fig:success_math_2} shows CoDR untangling a distorted algebraic substitution (introducing a sign and term error) and re-evaluating the true algebraic path ($P=50$).


\subsection{Regression Cases}
Despite its overall gains, CoDR is not monotonic and can occasionally degrade a correct base output. Tracking its failure modes provides sharp boundaries regarding the limitations of training-free inference-time interventions.

Figures~\ref{fig:regress_code} and \ref{fig:regress_math} illustrate the two primary modes of regression: dropped context and fresh hallucinations. In Figure~\ref{fig:regress_code}, CoDR correctly isolates a cluster of low-confidence stylistic tokens at the block boundary but accidentally drops a vital preceding string operation (\texttt{split()}) during the unmask-refill process, reducing list evaluation to character-level iterations. In Figure~\ref{fig:regress_math}, the base sampler actually derives the exact combinatorial solution. However, CoDR triggers a stylistic layout change, and during generation, the underlying MDLM introduces a new arithmetic corruption ($\binom{14}{5} \to 1002$).

\paragraph{Analysis}
Our diagnostic inspection reveals that regressions are fundamentally driven by \emph{spurious variations during the refill pass} rather than systematic failures in drift detection. Because the targeted tokens are fed directly back into the underlying MDLM's standard conditional sampling loop, the refinement step remains statistically bounded by the base model's inherent local calculation failure rates. This explains why code generation tasks exhibit virtually zero regressions (e.g., 0 out of 164 on HumanEval), while arithmetic tasks are more vulnerable to fresh hallucinations.

\section{AI Ethics Statement}
The authors acknowledge the use of LLM assistants only for context refinement and some code refinement during the preparation of this manuscript, and have checked carefully for any potential ethical concerns.

\begin{figure*}[htbp]
\centering
\small
\begin{minipage}{0.49\textwidth}
\hrule \vspace{2pt}
\textbf{Base Sampler (HumanEval/89) -- FAIL} \\
\vspace{-6pt} \hrule \vspace{4pt}
\begin{lstlisting}[language=Python, basicstyle=\ttfamily\scriptsize, breaklines=true]
def encrypt(s):
    encrypted_string = ''
    for char in s:
        if char.isalpha():
            # BUG: malformed shift under sparse context
            encrypted_string += chr((ord(char) - ord('a') * 2) % 26 + ord('a'))
        else:
            encrypted_string += char
    return encrypted_string
\end{lstlisting}
\end{minipage}
\hfill
\begin{minipage}{0.49\textwidth}
\hrule \vspace{2pt}
\textbf{CoDR Refinement (HumanEval/89) -- PASS} \\
\vspace{-6pt} \hrule \vspace{4pt}
\begin{lstlisting}[language=Python, basicstyle=\ttfamily\scriptsize, breaklines=true]
def encrypt(s):
    encrypted_string = ''
    for char in s:
        if char.islower():
            # FIXED: Correct +4 rotation logic refilled
            encrypted_string += chr((ord(char) - ord('a') + 4) % 26 + ord('a'))
        else:
            encrypted_string += char
    return encrypted_string
\end{lstlisting}
\end{minipage}
\caption{Code Success 1: The base sampler commits to operator-precedence garbage (\texttt{ord('a')*2}). CoDR isolates the drift at this precise expression and recovers the exact rotation algorithm under dense context.}
\label{fig:success_code_1}
\end{figure*}

\begin{figure*}[htbp]
\centering
\small
\begin{minipage}{0.49\textwidth}
\hrule \vspace{2pt}
\textbf{Base Sampler (HumanEval/133) -- FAIL} \\
\vspace{-6pt} \hrule \vspace{4pt}
\begin{lstlisting}[language=Python, basicstyle=\ttfamily\scriptsize, breaklines=true]
def sum_squares(lst):
    # BUG: math library invoked without being imported
    rounded_lst = [math.ceil(x) for x in lst]
    squared_lst = [x ** 2 for x in rounded_lst]
    return sum(squared_lst)
\end{lstlisting}
\end{minipage}
\hfill
\begin{minipage}{0.49\textwidth}
\hrule \vspace{2pt}
\textbf{CoDR Refinement (HumanEval/133) -- PASS} \\
\vspace{-6pt} \hrule \vspace{4pt}
\begin{lstlisting}[language=Python, basicstyle=\ttfamily\scriptsize, breaklines=true]
import math                                     # FIXED: Prepended during refill
def sum_squares(lst):
    rounded_lst = [math.ceil(x) for x in lst]
    squared_lst = [x ** 2 for x in rounded_lst]
    return sum(squared_lst)
\end{lstlisting}
\end{minipage}
\caption{Code Success 2: Confidence drift captures a structural failure where the base model utilizes a library function but omits the mandatory import header, leading to a runtime \texttt{NameError}. CoDR repairs the header seamlessly.}
\label{fig:success_code_2}
\end{figure*}

\begin{figure*}[htbp]
\centering
\small
\begin{minipage}{0.49\textwidth}
\hrule \vspace{2pt}
\textbf{Base Sampler (GSM8K/50) -- WRONG} \\
\vspace{-6pt} \hrule \vspace{4pt}
Lloyd's chickens produce 252 eggs per day. There are 7 days in a week, so: $252 \times 7 = 1764 \text{ eggs/week}$. There are 12 eggs in a dozen, so:
\[ \frac{1764 \text{ eggs/week}}{12 \text{ eggs/dozen}} = 142 \text{ dozens/week} \] \hfill \textbf{\color{purple}(BUG: Arithmetic Slip)} \\
Lloyd sells each dozen for \$2, so:
\[ 142 \text{ dozens/week} \times 2 \text{ dollars/dozen} = 284 \text{ dollars/week} \]
Therefore, Lloyd makes $\boxed{284}$.
\end{minipage}
\hfill
\begin{minipage}{0.49\textwidth}
\hrule \vspace{2pt}
\textbf{CoDR Refinement (GSM8K/50) -- CORRECT} \\
\vspace{-6pt} \hrule \vspace{4pt}
Lloyd's chickens produce 252 eggs per day. There are 7 days in a week, so: $252 \times 7 = 1764 \text{ eggs/week}$. There are 12 eggs in a dozen, so:
\[ \frac{1764 \text{ eggs/week}}{12 \text{ eggs/dozen}} = 147 \text{ dozens/week} \] \hfill \textbf{\color{teal}(FIXED: Long Division)} \\
Lloyd sells each dozen for \$2, so:
\[ 147 \text{ dozens/week} \times 2 \text{ dollars/dozen} = 294 \text{ dollars/week} \]
Therefore, Lloyd makes $\boxed{294}$.
\end{minipage}
\caption{Math Success 1: The base model suffers an arithmetic calculation failure ($1764/12=142$). CoDR re-scores the fractional setup and enforces the mathematically sound quotient ($147$), altering the downstream revenue projection.}
\label{fig:success_math_1}
\end{figure*}

\begin{figure*}[htbp]
\centering
\small
\begin{minipage}{0.49\textwidth}
\hrule \vspace{2pt}
\textbf{Base Sampler (GSM8K/931) -- WRONG} \\
\vspace{-6pt} \hrule \vspace{4pt}
Equations: (1) $P+P+J=100$, (2) $P=J+10$, (3) $P=P+J$. From the third equation, substitute $P$ in the first: $2P+J=100$. \\
Now, substitute $P$ from the second equation:
\[ 2P + (P - 10) = 100 \] \hfill \textbf{\color{purple}(BUG: Distorted Substitution)} \\
$3P - 10 = 100 \implies P = 30$. So, Peter is $30$ years old.
\end{minipage}
\hfill
\begin{minipage}{0.49\textwidth}
\hrule \vspace{2pt}
\textbf{CoDR Refinement (GSM8K/931) -- CORRECT} \\
\vspace{-6pt} \hrule \vspace{4pt}
Relationships: (1) $P+P+J=100$, (2) $P=J+10$, (3) $P=P+J$. Using the third equation, substitute $P$ in the first: $2P+J=100$. \\
Now, using the second equation, substitute $P$ in the above equation:
\[ 2P + 0 + 10 = 100 \] \hfill \textbf{\color{teal}(FIXED: Algebraic Soundness)} \\
$2P = 100 \implies P = 50$. So, Peter is $50$ years old.
\end{minipage}
\caption{Math Success 2: The base model maps the constraints accurately but introduces a sign and term error during variable evaluation. CoDR tracks the logical fracture and re-evaluates the true algebraic path ($P=50$).}
\label{fig:success_math_2}
\end{figure*}

\begin{figure*}[htbp]
\centering
\small
\begin{minipage}{0.49\textwidth}
\hrule \vspace{2pt}
\textbf{Base Sampler (MBPP/117) -- PASS} \\
\vspace{-6pt} \hrule \vspace{4pt}
\begin{lstlisting}[language=Python, basicstyle=\ttfamily\scriptsize, breaklines=true]
def long_words(n, words):
    words = words.split()     # Crucial string-to-list setup
    long_words = [word for word in words if len(word) > n]
    return long_words
\end{lstlisting}
\end{minipage}
\hfill
\begin{minipage}{0.49\textwidth}
\hrule \vspace{2pt}
\textbf{CoDR Refinement (MBPP/117) -- FAIL} \\
\vspace{-6pt} \hrule \vspace{4pt}
\begin{lstlisting}[language=Python, basicstyle=\ttfamily\scriptsize, breaklines=true]
def long_words(n, words):
    # REGRESSION: split() operation dropped during refill
    return [word for word in words if len(word) > n]
\end{lstlisting}
\end{minipage}
\caption{Code Regression: CoDR correctly isolates a cluster of low-confidence stylistic tokens at the block boundary but accidentally drops a vital preceding token string operation (\texttt{split()}) during the unmask-refill process, reducing list evaluation to character-level iterations.}
\label{fig:regress_code}
\end{figure*}

\begin{figure*}[htbp]
\centering
\small
\begin{minipage}{0.49\textwidth}
\hrule \vspace{2pt}
\textbf{Base Sampler (MATH-500 Doc 330) -- PASS} \\
\vspace{-6pt} \hrule \vspace{4pt}
1. Calculate the total number of ways to choose 5 members out of 14:
\[ \binom{14}{5} = \frac{14!}{5!(14-5)!} = 2002 \] \hfill \textbf{\color{teal}(CORRECT Combinatorics)} \\
2. Ways from Republicans: $\binom{8}{5} = 56$. 3. Ways from Democrats: $\binom{6}{5} = 6$. \\
4. Subtract exclusions: $2002 - 56 - 6 = 1940$.
\end{minipage}
\hfill
\begin{minipage}{0.49\textwidth}
\hrule \vspace{2pt}
\textbf{CoDR Refinement (MATH-500 Doc 330) -- FAIL} \\
\vspace{-6pt} \hrule \vspace{4pt}
1. Calculate the total number of ways to choose 5 members out of 14:
\[ \binom{14}{5} = \frac{14!}{5!(14-5)!} = 1002 \] \hfill \textbf{\color{purple}(REGRESSION: Hallucinated Factorial)} \\
2. Ways from Republicans: $\binom{8}{5} = 56$. 3. Ways from Democrats: $\binom{6}{5} = 6$. \\
4. Subtract exclusions: $1002 - 56 - 6 = 940$.
\end{minipage}
\caption{Math Regression: A classic instantiation of fresh hallucination during resampling. The base sampler derives the exact combinatorial solution. CoDR triggers a stylistic layout change, but the model introduces a new arithmetic corruption ($\binom{14}{5} \to 1002$) during generation.}
\label{fig:regress_math}
\end{figure*}

\end{document}